\documentclass[runningheads]{llncs}
\usepackage{graphicx}

\usepackage{tikz}
\usepackage{comment}
\usepackage{amsmath,amssymb} 
\usepackage{color}
\usepackage{booktabs}
\usepackage{hyperref}
\usepackage[accsupp]{axessibility}  

\hypersetup{
    colorlinks=true,
    filecolor=magenta,      
    pdfpagemode=FullScreen,
    }

\def\ie{\textit{i.e.}}

\begin{document}
\pagestyle{headings}
\mainmatter
\def\ECCVSubNumber{100}  

\title{Egocentric Video Task Translation \\ @ Ego4D Challenge 2022} 

\titlerunning{Egocentric Video Task Translation}
%
\author{Zihui Xue\inst{1,2} \and
Yale Song\inst{2} \and
Kristen Grauman \inst{1,2} \and
Lorenzo Torresani \inst{2}}
\authorrunning{Z. Xue et al.}
%
\institute{The University of Texas at Austin \and Meta AI \\
\email{sherryxue@utexas.edu,\{yalesong, grauman, torresani\}@meta.com}}
\maketitle

\begin{abstract}
This technical report describes the EgoTask Translation approach that explores relations among a set of egocentric video tasks in the Ego4D challenge. To improve the primary task of interest, we propose to leverage existing models developed for other related tasks and design a task translator that learns to ``translate'' auxiliary task features to the primary task. With no modification to the baseline architectures, our proposed approach achieves competitive performance on two Ego4D challenges, ranking the 1st in the talking to me challenge and the 3rd in the PNR keyframe localization challenge.

\end{abstract}

\section{Introduction}
While video understanding in the third-person setting has focused overwhelmingly on the single task of action recognition, the recently released large-scale egocentric dataset, Ego4D~\cite{ego4d}, aims at capturing the multifaceted aspects of human-human and human-object interactions. The Ego4D challenge consists of a diverse set of spatiotemporal tasks that extend much beyond action categorization. Examples include the talking to me, looking at me challenge for human-human interactions and PNR keyframe localization, long-term action anticipation challenge for human-object interactions.

It is apparent that strong synergies exist among these tasks. For instance, identifying whether someone in the scene is talking to the camera wearer (\ie, talking to me challenge) is closely related to the task of identifying whether someone is looking at the camera wearer (\ie, looking at me challenge). Identifying the presence or absence of an object state change in the video (\ie, object state change classification challenge) can provide useful hints for the task of PNR keyframe localization. Motivated by such finding, we aim to model the relations among different tasks and propose a general solution to improve individual task performance with the assistance of related tasks.

In order to study task relations, we propose to leverage existing baseline models developed for each challenge and design a task translator to ``translate'' features produced by these task-specific models for improving the task of interest. The proposed EgoTask Translation framework can incorporate heterogeneous video models categorized for each challenge, and thus offers a general and flexible solution orthogonal to model architecture improvements. 

\section{Approach}
Given a set of $K$ tasks, we assume that each task is associated with its own dataset. Let the dataset for task $\mathcal T_k$ be $\{(\mathbf x_i^{\mathcal T_k}, y_i^{\mathcal T_k})\}_{i=1}^{N_k}$, where $(\mathbf x_i^{\mathcal T_k}, y_i^{\mathcal T_k})$ denotes the $i$-th pair of (input video, output label) and $N_k$ represents dataset length.  The objective is to improve the primary task $\mathcal T_p$ with the assistance of the other $K-1$ auxiliary tasks. 

We propose a two-stage training framework. In the first stage, a task-specific model is trained on each individual task from raw audiovisual inputs. Let $f_k$ denote the task-specific model for task $\mathcal T_k$. This step allows each model to be optimized with respect to the individual task. Note that unlike previous approaches that study visual task relations~\cite{zamir2018taskonomy}, a unified design across tasks is not required, thus we can resort to different baseline models developed for each challenge to use within our framework.  

In the second stage, we design a task translator that takes features produced by task-specific models as input and outputs predictions for the primary task. Formally, let $\mathbf h_k \in \mathbb{R}^{T_k \times D_k}$ be features produced by the $k$-th task-specific model $f_k$, where $T_k$ corresponds to the temporal dimension and $D_k$ denotes the per-frame feature dimension for model $f_k$. Following the feature extraction step, we design a projection layer $\mathbf P_k\in \mathbb{R}^{D_k \times D}$ for each $f_k$ to map task-specific features to a shared latent feature space. The projected features are then concatenated together along the temporal dimension to obtain a sequence of task-specific tokens. To retain task and positional information in $\mathbf h_i$, we add \emph{task positional embeddings} $\mathbf{P}_{task} \in \mathbb{R}^{\sum_{k=1}^K T_k \times D}$ to the concatenated features. A transformer encoder~\cite{vaswani2017attention} is then adopted to capture inter-task and inter-frame relations among features. It consists of $L$ transformer layers, and we denote the propagation rule of each layer by $\mathbf z^{l+1} = Transformer(\mathbf z^l)$. Finally, we adopt a decoder head $Decoder^{\mathcal T_p}$ to obtain predictions for the primary task $\mathcal T_p$. In all, there are four major steps: (1) feature extraction; (2) feature projection; (3) transformer fusion; and (4) feature decoding. The propagation rule is summarized below:
\begin{align}
    \mathbf h_k & = f_k(\mathbf x_i^{\mathcal T_p}),\quad \forall k\in \{1,2,\cdots, K\} \label{eq.feat}\\
    \mathbf z^0 & = [\mathbf P_1 \mathbf h_1, \mathbf P_2 \mathbf h_2, \cdots, \mathbf P_K \mathbf h_K] + \mathbf{P}_{task} \\
    \mathbf z^{l+1} & = Transformer(\mathbf z^l), \quad \forall l\in \{0,1,\cdots, L-1\} \\
    y_{pred_i}^{\mathcal T_p} & = Decoder^{\mathcal T_p}(\mathbf z^L)
\end{align}

\begin{figure}[t]
  \centering
   \includegraphics[width=0.7\linewidth]{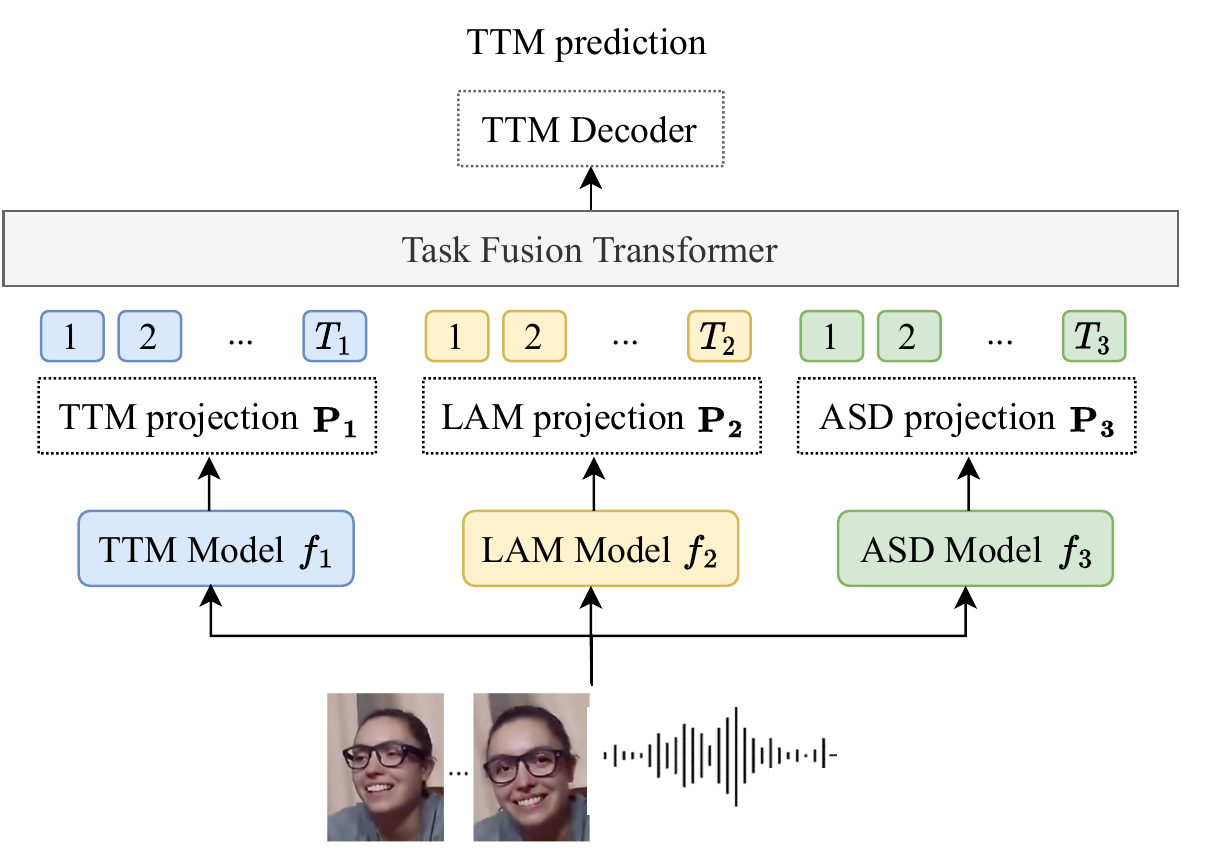}
   \caption{An illustration of our proposed task translator, where the primary task is talking to me (TTM), and auxiliary tasks are looking at me (LAM) and active speaker detection (ASD). In the first stage, we train three task-specific models $f_1$, $f_2$ and $f_3$ with respect to each task. In the second-stage of training, the task translator learns to ``translate'' features produced by task-specific models into TTM predictions.}
   \label{fig.framework}
\end{figure}

\textbf{Remarks}. (1) During the second-stage training, we freeze the task-specific models and optimize the task translator with respect to the primary task dataset $\{\mathbf x_i^{\mathcal T_p}, y_i^{\mathcal T_p}\}_i$. $y_{pred_i}^{\mathcal T_p}$ denotes the prediction on the $i$-th sample of the primary task dataset $\mathcal T_p$ given by the task translator. (2) Equation \ref{eq.feat} is a simplification of the feature extraction process. In fact, there is one additional step to transform the video input $\mathbf x_i^{\mathcal T_p}$ to be consistent with the input requirement of the task-specific model. For instance, if the $k$-th task-specific model $f_k$ is trained on 8-second video clips at 2 frames per seconds (FPS), and the primary task dataset provides 16-second videos at 4 FPS, we first subsample a video from ${\mathcal T_p}$ to match the FPS and slide $f_k$ in a moving window to extract features for the 16-second video, where the window size is the auxiliary task video duration (\ie, 8 seconds in this example) and the stride size is a hyperparameter. This design allows our model to be applied to multiple video datasets with different frame rates and temporal spans, achieving maximum flexibility. Finally, Figure \ref{fig.framework} provides an illustration of our approach on the talking to me challenge.

\section{Experiments}
\subsection{Experimental Setup}
 We evaluate the proposed EgoTask Translation approach on three different setups: (1) looking at me and active speaker detection are considered as auxiliary tasks to improve talking to me; (2) object state change classification is the auxiliary task and PNR keyframe localization is the primary task; (3) action recognition is adopted as the auxiliary task and long-term action anticipation is the primary task. For all the tasks involved, we adopt baseline models provided in the Ego4D challenge\footnote{We use model checkpoints provided in \url{https://github.com/EGO4D}, or follow the training script for a baseline model if checkpoints are not available for some task.} as task-specific models.
\begin{table}[!t]
\begin{center}
\caption{Results of talking to me challenge}
\label{table.ttm}
\begin{tabular}{lcc}
\toprule
Method & Accuracy (\%) & mAP (\%)\\
\midrule
Random Guess~\cite{ego4d} & 47.41 & 50.16\\
ResNet-18 Bi-LSTM~\cite{ego4d} & 49.75 & 55.06\\
EgoTask Translation (ours) & \textbf{55.93} & \textbf{57.52}\\
\bottomrule
\end{tabular}
\end{center}
\end{table}

\begin{table}[!t]
\begin{center}
\caption{Results of PNR keyframe localization challenge. `loc. error' denotes temporal localization error (seconds). Lower is better}
\label{table.pnr}
\begin{tabular}{lcc}
\toprule
Method & loc. error\\
\midrule
Always Center Frame~\cite{ego4d} & 1.056\\
I3D ResNet-50~\cite{ego4d} & 0.755\\
Video Swin Transformer~\cite{escobar2022video} & 0.660 \\
SViT~\cite{ben2022structured} & 0.660 \\
EgoTask Translation (ours) & \textbf{0.655} \\
\bottomrule
\end{tabular}
\end{center}
\end{table}

\begin{table}[!t]
\begin{center}
\caption{Results of long-term action anticipation challenge. `ED@(Z=20)' denotes the edit distance at 20 future time stamps. Lower is better}
\label{table.lta}
\begin{tabular}{lccc}
\toprule
Method & \multicolumn{3}{c}{ED@(Z=20)}\\
 & Verb & Noun & Action\\
\midrule
SlowFast-Transformer~\cite{ego4d} & 0.74 & 0.78 & 0.94 \\
Video + CLIP ~\cite{das2022video+} & 0.74 & 0.77 & 0.94 \\
Hierarchical Multitask MLP Mixer~\cite{mascaro2022intention} & 0.74 & \textbf{0.74} & \textbf{0.93} \\
EgoTask Translation (ours) & \textbf{0.72} & 0.76 & \textbf{0.93} \\
\bottomrule
\end{tabular}
\end{center}
\end{table}

\subsection{Results}
Table \ref{table.ttm}-\ref{table.lta} presents our results on test datasets for the three tasks. We observe consistent performance gain brought by our task translator. For instance, on talking to me challenge, our approach leads to +6.18\% test accuracy improvement when compared with the baseline model (\ie, ResNet-18 Bi-LSTM). For the PNR keyframe localization task, the task translator incorporates I3D ResNet-50 as task-specific models, yet can outperform more advanced backbone architectures such as Video Swin Transformer. These results demonstrate the efficiency and general applicability of the proposed task translator.


\section{Conclusion and Limitations}
We propose EgoTask Translation, a general and flexible framework for learning with multiple egocentric video tasks. The limitation lies in that the task translator is specially designed for one primary task, and changing the primary task will require retraining of the task translator. In the future, we plan to extend the task translator to be optimized for multiple tasks simultaneously. 
\clearpage
%
%
\bibliographystyle{splncs04}
\bibliography{egbib}
\end{document}